\documentclass[runningheads]{llncs}
\usepackage{graphicx}

\usepackage{times}
\usepackage{xcolor}
\usepackage{soul}
\usepackage[utf8]{inputenc}
\usepackage[small]{caption}

\usepackage{times}
\usepackage{epsfig}
\usepackage{graphicx}
\usepackage{amsmath}
\usepackage{amssymb}
\usepackage{arydshln}
\usepackage{colortbl}
\usepackage{color}
\usepackage{float}
\usepackage{caption}

\begin{document}
\title{On Minimum Discrepancy Estimation for Deep Domain Adaptation}
%
%\titlerunning{Abbreviated paper title}
% If the paper title is too long for the running head, you can set
% an abbreviated paper title here
%
\author{Mohammad Mahfujur Rahman\inst{1} \and
Clinton Fookes\inst{1} \and
Mahsa Baktashmotlagh \inst{1} \and
Sridha Sridharan\inst{1}}
\authorrunning{M. M. Rahman et al.}
% First names are abbreviated in the running head.
% If there are more than two authors, 'et al.' is used.
%
\institute{Image and Video Laboratory, Queensland University of Technology (QUT), \\ Brisbane, QLD, Australia \\
%\email{lncs@springer.com}\\
%\url{http://www.springer.com/gp/computer-science/lncs} \and
%ABC Institute, Rupert-Karls-University Heidelberg, Heidelberg, Germany\\
\email{\{m27.rahman, c.fookes, 
m.baktashmotlagh, s.sridharan\}@qut.edu.au}}
\maketitle              % typeset the header of the contribution
\begin{abstract}
In the presence of large sets of labeled data, Deep Learning (DL) has accomplished extraordinary triumphs in the avenue of computer vision, particularly in object classification and recognition tasks. However, DL cannot always perform well when the training and testing images come from different distributions or in the presence of domain shift between training and testing images. They also suffer in the absence of labeled input data. Domain adaptation (DA) methods have been proposed to make up the poor performance due to domain shift. In this paper, we present a new unsupervised deep domain adaptation method based on the alignment of second order statistics (covariances) as well as maximum mean discrepancy of the source and target data with a two stream Convolutional Neural Network (CNN). We demonstrate the ability of the proposed approach to achieve state-of-the-art performance for image classification on three benchmark domain adaptation datasets: Office-31 \cite{Saenko:2010:AVC:1888089.1888106}, Office-Home \cite{venkateswara2017Deep} and Office-Caltech \cite{DBLP:conf/cvpr/GongSSG12}.

\end{abstract}
%
%

%%%%%%%%% BODY TEXT
\section{Introduction}

Deep Neural Networks (DNN) \cite{0483bd9444a348c8b59d54a190839ec9} have brought tremendous advances across many machine learning tasks and applications such as object detection \cite{Girshick:2014:RFH:2679600.2679851}, object recognition \cite{NIPS2012_4824}, speech recognition \cite{6639346}, person re-identification \cite{8354250} and machine translation \cite{DBLP:journals/corr/SutskeverVL14}. For an example, in \cite{DBLP:journals/corr/GoodfellowBIAS13} a DNN achieves 97.84\% accuracy in multi digit number classification from street view images because of the ability of joint feature and classifier learning of the DNN. The dramatic success of large scale image classification based on DNNs commenced in 2012. In \cite{NIPS2012_4824}, they  attained the best performance in the ImageNet Large Scale Visual Recognition Challenge (ILSVRC) by developing AlexNet. These victories were achieved in part from the accessibility of large labeled datasets such as the widely used ImageNet \cite{NIPS2012_4824}. While the introduction of such datasets have unlocked many breakthroughs, the process of obtaining such labels still remains a time consuming and manual task.

In object recognition or classification, the training images may be different than the target images due to backgrounds, camera viewpoints, object transformations and human selection preference. When the source data and target data distributions are dissimilar, classifier's performance can be significantly impacted. In computer vision, this is generally known as dataset bias or dataset shift \cite{5995347,DBLP:journals/corr/Long0J16}. Learning a discriminative model of different distributions of training and test data is known as domain adaptation \cite{5288526,7078994,YANG2018615}. The principle objective of unsupervised domain adaptation algorithms is to interface the source and target distributions by acquiring a domain-constant informations where the target data are used without any labels.

Recent investigations have demonstrated that deep neural networks learn more transferable components for unsupervised domain adaptation \cite{coral}. Recently, unsupervised domain adaptation methods \cite{dcoral,coral,Hong_2018_CVPR,Hu_2018_CVPR,Saito_2018_CVPR,Mancini_2018_CVPR,Murez_2018_CVPR,Volpi_2018_CVPR,Pinheiro_2018_CVPR} have been proposed where features are adapted by aligning the second order statistics of the source and target data. Although \cite{coral} introduces a new loss named Correlation Alignment (CORAL) Loss, it depends on a linear transformation, and it is not an end-to-end trainable method. After feature extraction, the linear transformation is applied, and a Support Vector Machine (SVM) classifier is trained in another phase. Moreover, the features are fixed in these type of shallow domain adaptation methods. The approach in \cite{coral} is extended in \cite{dcoral} to incorporate the CORAL loss directly into deep neural networks. Maximum Mean Discrepancy (MMD) is another popular metric for feature adaptation. MMD based DA techniques have achieved great success to minimize the discrepancy between source and target data. MMD can also be incorporated with deep neural networks to achieve stronger performance over conventional methods.

In our approach, we get motivation from both of the above top performing metrics and propose a new domain adaptation method which leverages the advantages of both feature adaptation metrics: CORAL and MMD. The difference between previous research and our work is that previous approaches either minimize the source and target data discrepancy using maximum mean discrepancy or second order statistics for feature adaptation. However, in our approach we minimize the discrepancy using both metrics (MMD and CORAL) for feature adaptation. MMD based methods for domain adaptation utilize symmetric transformation to distributions of the source and target data whereas CORAL based approaches apply asymmetric transformation. However, symmetric transformations neglects the dissimilarities between the source
and target data. On the other hand, asymmetric transformations attempt to link the source and target domains \cite{DBLP:journals/corr/SunFS16}. CORAL aligns the second order statistics that can be reconstructed utilizing all eigenvectors and eigenvalues instead of aligning only the top $k$ eigenvectors and eigenvalues as subspace based methods \cite{6751479}.

We present an assessment of our proposed deep domain adaptation by aligning covariances or second order statistics and maximum mean discrepancy within a two stream of CNN on three benchmark datasets: Office-31 \cite{Saenko:2010:AVC:1888089.1888106}, the recently released Office-Home \cite{venkateswara2017Deep} and Office-Caltech \cite{DBLP:conf/cvpr/GongSSG12}.

In summary, the contributions of this paper are given as follows:
%\begin{enumerate}
\begin{itemize}
\item We propose a novel deep neural network approach for unsupervised domain adaptation in the context of image classification in computer vision.

\item The proposed deep domain adaptation architecture jointly adapts features using two popular feature adaptation metrics: MMD and CORAL. 

\item We report competitive  accuracy  compared  to  the  state-
of-art  methods on three benchmark domain adaptation datasets for image classification. We achieve the best average image classification accuracies on three datasets compared to other state-of-the art methods.

%\item We visualize the extracted Fully Connected-7 (fc-7) CNN features using t-Distributed Stochastic Neighbor Embedding (t-SNE) \cite{ictdbid:2777} to visualize the effectiveness of our model.

\end{itemize}

The rest of the paper is organized as follows: Section 2 describes related research; the proposed methodology is described in Section 3; Section 4 illustrates a comprehensive evaluation; and finally, Section 5 concludes the paper. 

\section{Related Works}

There have been many domain adaptation methods \cite{Choi_2018_CVPR,Pinheiro_2018_CVPR,Volpi_2018_CVPR,Murez_2018_CVPR,Mancini_2018_CVPR,Saito_2018_CVPR,DBLP:journals/corr/TzengHDS15,dcoral} proposed in recent years to solve the problem of domain bias. All the methods can be categorized into two main categories, Conventional Domain Adaptation and Deep Domain Adaptation methods. The conventional domain adaptation methods develop their model into two stages, feature extraction and classification. In the first phase, these domain adaptation methods extract features and in the second phase, a classifiers is trained to classify the objects. However, the performance of these DA methods are not satisfactory. 

Obtaining the features using deep neural network even without adaptation technique outperform the conventional DA methods by large margin. However, the results achieved with the Deep Convolutional Activation Features (DeCAF) \cite{Donahue:2014:DDC:3044805.3044879} even without using any adaptation technique to the target data are remarkably better than the outcomes acquired with any conventional domain adaptation methods because DNNs extract more robust features using nonlinear transform. As a result deep neural network based domain adaptation methods are getting popular day by day.

MMD is a popular metric for measuring the distributions of source and target samples. Tzeng et al. \cite{DBLP:journals/corr/TzengHZSD14} proposed the Deep Domain Confusion (DDC) domain adaptation framework based on a confusion layer for the discrepancy between source and target data. In \cite{DBLP:journals/corr/TzengHDS15}, the previous work is extended by introducing soft label distribution matching loss. Long et al. \cite{DBLP:journals/corr/Long015} proposed the Domain Adaptation Network (DAN) that propose the integration of MMDs defined among several layers, including the soft prediction layer. This idea was further improved by introducing residual transfer networks \cite{DBLP:journals/corr/Long0J16} and Joint Adaptation Networks \cite{DBLP:journals/corr/Long0J16a}. Venkateswara et al. \cite{venkateswara2017Deep} proposed a new Deep Hasing Network for unsupervised domain adaptation where hash codes are used to address the domain adaptation issue.

Another popular metric for feature adaptation between domains is aligning covariance or second order statistics which is known as Correlation Alignment. In \cite{coral,dcoral}, unsupervised deep domain adaptation techniques have been proposed where domain shift is minimized by aligning the covariances of the source and target data. The idea is similar to Deep Domain Confusion (DDC) \cite{DBLP:journals/corr/TzengHZSD14} and Deep Adaptation Network (DAN) \cite{DBLP:journals/corr/Long015} except that the CORAL loss is used instead of MMD to minimize the discrepancy between source and target data. Both \cite{coral,dcoral} introduces a new loss named coral loss which is the distance between the second-order statistics of the source and target representations. In \cite{DBLP:journals/corr/KoniuszTP16}, a deep domain adaptation approach based on the mixture of alignments of second order or higher-order scatter statistics between source and target distributions has been proposed. All these methods utilized two stream of CNN where the source network and target network combined at the classifier level. Another deep domain adaptation method is Domain-Adversarial Neural Networks (DANN) \cite{Ganin:2016:DTN:2946645.2946704} which introduces a new deep learning domain adaptation approach by integrating a gradient reversal layer into the standard architecture. This gradient reversal layer do not change during forward propagation, but during back propagation its gradient reverse.

In our work, we adapt the features using both CORAL and MMD metric to minimize the dissimilarity between the source and target domains. CORAL is used to align the second order statistics and MMD is used to align higher order statistics.

\section{Proposed Approach }

\begin{figure*}
\begin{center}
\includegraphics[width=1.0\linewidth]{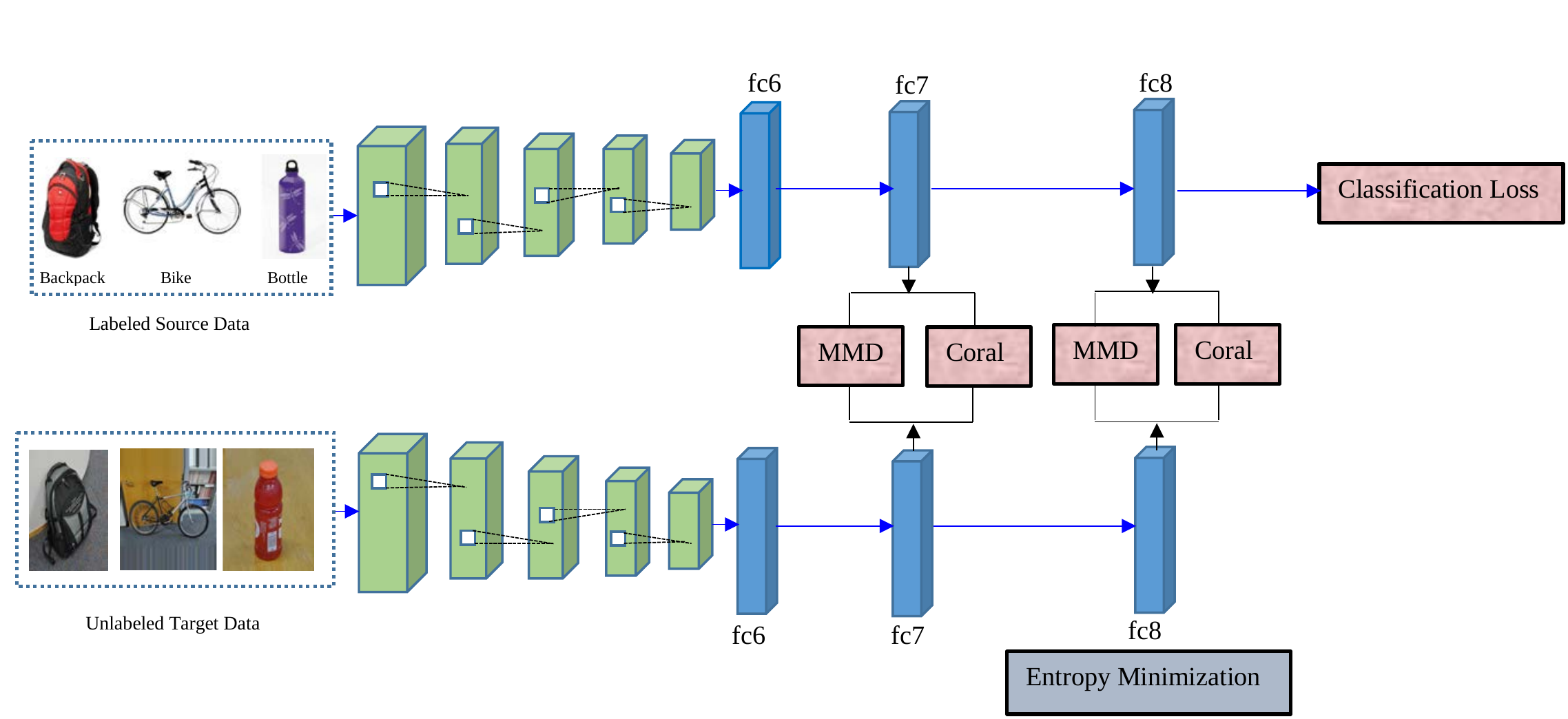}
\end{center}
   \caption{Overview of our proposed methodology. The classifiers and features of both source and target data are adapted simultaneously. MMD and CORAL loss layers are added in fc7 and fc8 layers of two stream of CNN. CORAL layer align the second order statistics and MMD layer aligns the higher order statistics.}
\label{fig:twocol}
\end{figure*}

Our proposed methodology is illustrated in Figure \ref{fig:twocol}. In Our proposed method, the features of the source and target domains are jointly adapted using CORAL and MMD metrics. The source and target data uses two separate CNNs. In fc7 and fc8 layers, CORAL and MMD loss layer are added to minimize the discrepancy between the source and target data. Finally, the discrepancy between source and target data is minimized by entropy minimization of the unlabeled target data.

We consider the unsupervised domain adaptation scenario where labeled source data and unlabeled target data are available. Let us consider that the source domain data samples are $D_s=\{X_i^s\}$ with available labels $L_s=\{Y_i\}$ and the target data samples are $D_t = \{X_i^t\}$ without labels. The number of source and target samples are $N_s$ and $N_t$ respectively. Let the classifiers for source domain and target domain be $F_s(X_i^s)$ and $F_t(X_i^t)$ respectively. The distribution of the data of source and target domains are non-identical, i.e., $P_s(X_i^s,Y_s)$ $\neq$ $P_t(X_i^t,Y_t)$. We build a deep learning architecture which aids the learning of a transfer classifiers, such as $Y = F_s(X_i^s) = F_t(X_i^t)$ to minimize the source-target discrepancy or mismatch.

We propose a new deep DA method which has two streams of convolutional Neural Network (CNN), one for source data and another for target data. It adapts features by aligning second order statistics and maximum mean discrepancy of the source and target data. The discrepancy of the source and target data are minimized by the following equation,

%begin{equation}
%min_{Fs=Ft } \frac{1}{Ns} \sum_{i=1}^{Ns} h(Fs(Xi),Yi) % \frac{1}{Nt}
%sum_{i=1}^{Nt} H (Ft (Xt)) + \min_{Fs,Ft} Dl (Ds, Dt)
%end{equation}

\begin{align}
\min_{F_s,F_t} D_l (D_s, D_t)_{fc7}+\min_{F_s,F_t} MMD^2(D_s,D_t)_{fc7} + \nonumber	\\   \min_{F_s,F_t} D_l (D_s, D_t)_{fc8}+\min_{F_s,F_t} MMD^2(D_s,D_t)_{fc8} +  \nonumber	\\ \sum_{i=1}^{N_t} H (F_t (X_i^t)).
\end{align}

Moreover, the proposed method also adapts the classifiers using entropy minimization. 

The features are adapted by aligning second order statistics as well as maximum mean discrepancy. We define the coral loss of the source and target activation features (such a loss function is used in prior work \cite{dcoral} ) as,

\begin{equation}
\min_{F_s,F_t} D_l (D_s, D_t) = \frac{1}{4d^2} \|C_s-C_t\|_F^2,
\end{equation}

where $C_s$ and $C_t$ denote the features covariance matrices of the source and target data and $||.||_F^2$ denotes the squared matrix Frobenius norm. The $C_s$ and $C_t$ are given by the following equation \cite{dcoral},

\begin{equation}
C_s = \frac{1}{N_s-1} (D_s^TD_s-\frac{1}{N_s}(1^TD_s)^T(1^TD_s) , 
\end{equation}
\begin{equation}
C_t = \frac{1}{N_t-1} (D_t^TD_t-\frac{1}{N_t}(1^TD_t)^T(1^TD_t).
\end{equation}

The features are further adapted by using another popular metric for feature adaptation, MMD. The MMD loss function is defined as,

\begin{align}
\min_{F_s,F_t} MMD^2(D_s,D_t) = \nonumber	\\ \| \frac{1}{N_s} \sum_{i=1}^{N_s} \phi (X_i^s) - \frac{1}{N_t} \sum_{i=1}^{N_t} \phi (X_i^t) \|_H^2,
\end{align}

where $\phi(X_i^s)$ denotes the feature map associated with kernel map,

 \begin{equation}
 K(X_i^s, X_i^t) =  <\phi (X_i^s), \phi (X_i^t)> K (X_i^s, X_i^t).
 \end{equation}

$K(X_i^s, X_i^t)$ is usually
defined as the convex combination of $L$ basis kernels $K_l(X_i^s, X_i^t)$ \cite{DBLP:journals/corr/YanDLWXZ17},

\begin{equation}
K(X_i^s, X_i^t) = \sum_{l=1}^{L} \beta_1 K_1 (X_i^s, X_i^t) s.t. \beta_1 \geq 0, \sum_{l=1}^{L} \beta_1 = 1.
\end{equation}

Since feature adaptation cannot eliminate the discrepancy\cite{DBLP:journals/corr/Long0J16}, we adapt classifiers along with feature adaptation. In this work, the classifier is  adapted by decreasing the entropy of class-conditional distribution on the target data $D_t$ (similar loss function has been proposed in prior work \cite{DBLP:journals/corr/Long0J16}),

\begin{equation}
\min_{F_t} \frac{1}{N_t} = \sum_{i=1}^{N_t} H (F_t (X_i^t)),
\end{equation}

where $H($·$)$ represents the class-conditional distribution entropy function.

\subsection{Discussion}

The main difference between our work and prior works is that they consider only one metric for feature adaptation whereas we consider two metrics for minimizing the discrepancy between the source and target data. In \cite{dcoral}, CORAL layer is used in between fc8 layers of the source and target CNNs, but we used CORAL layer in between fc7 and fc8 layers. It is mentioned that the MMD metric is used in between fc8 layers in \cite{DBLP:journals/corr/Long0J16} and MMD layer is used in between fc6, fc7 and fc8 layers in \cite{DBLP:journals/corr/Long015}. The difference between our work and \cite{DBLP:journals/corr/Long0J16} is that RTN uses Residual Transfer Network and MMD metric whereas we use simple AlexNet architecture that consists of 5 convolutional followed by 3 fully connected layers and CORAL and MMD metrics to adapt the features. In our research, we have found that if multiple feature adaptation metrics are used in between fc7 and fc8, we get better accuracy using simple CNN architecture, and the best configuration of domain adaptation architecture is to use feature adaptation metric in between
fc7 and fc8.

\begin{figure}
\begin{center}
\includegraphics[width=0.7\linewidth]{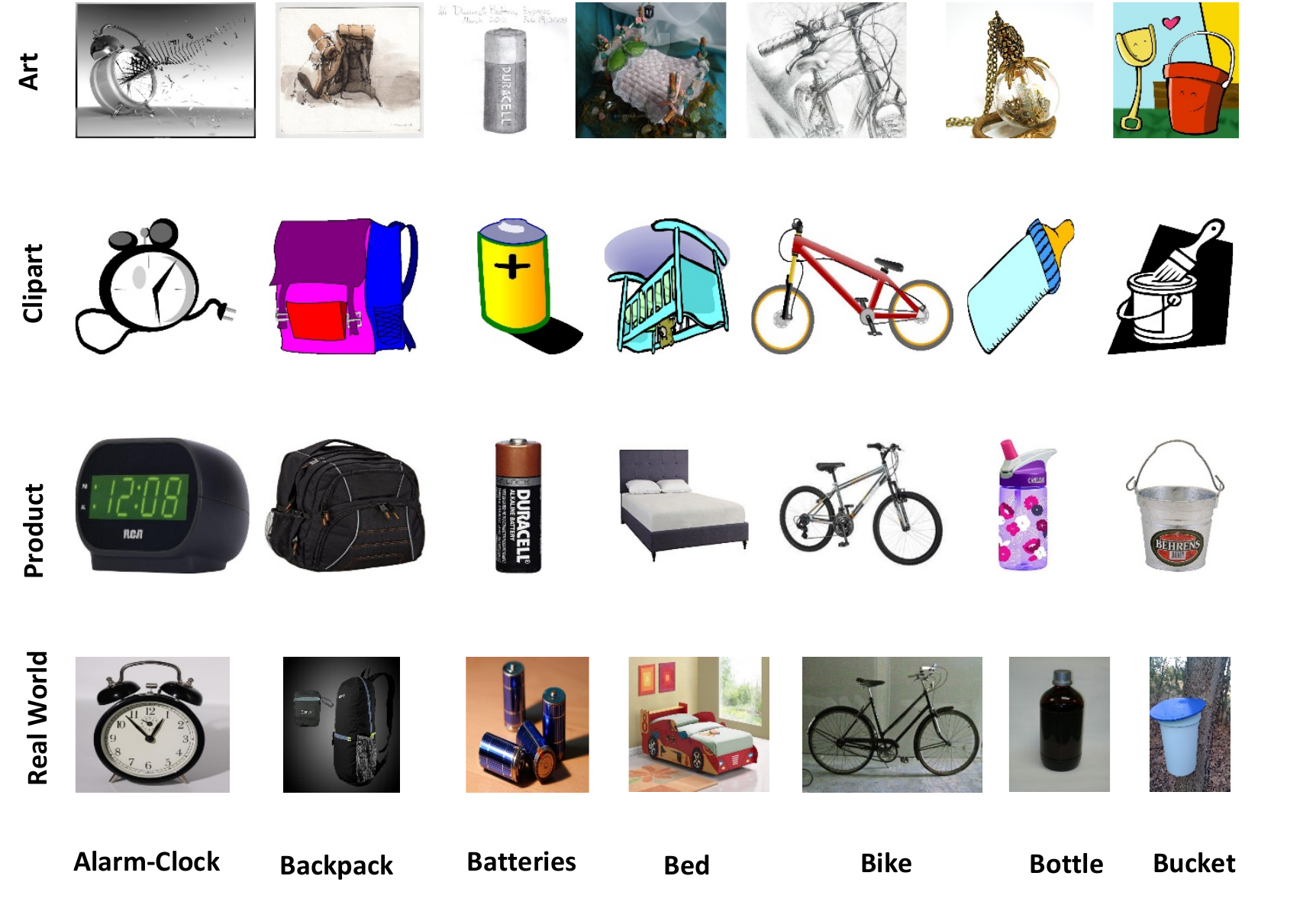}
\end{center}
   \caption{There are some example images that are taken from \textbf{Office-Home} dataset. It comprises of images of everyday objects. This dataset divided into 4 different domains; the \textbf{Clipart} domain comprises clipart images, the \textbf{Art} domain consists of sketches, paintings, artistic images, the \textbf{Product} domain comprises images which have no background and finally, the \textbf{Real-World} domain is created by taking images which are captured with a regular camera. The figure shows sample images from 7 of the 65 classes.}
\label{fig:office}
\end{figure}

\section{Experiments}

In this section we conduct extensive experiments to assess the proposed method and compare the method against recently published state-of-the-art unsupervised deep domain adaptation approaches.

\subsection{Datasets}
We evaluate all the methods on three standard domain adaptation benchmark datasets: Office-31 \cite{Saenko:2010:AVC:1888089.1888106}, Office-Home \cite{venkateswara2017Deep} and Office-Caltech \cite{DBLP:conf/cvpr/GongSSG12} in the context of image classification. 

\subsubsection{Office-31}

In the context of image classification, Office-31 is the most prominent benchmark dataset for domain adaptation. The dataset contains everyday object images from an office environment. It consists of 4110 images with 31 object categories and 3 image domains: \textbf{Amazon (A)} contains images downloaded from amazon.com, \textbf{DSLR (D)} contains images taken by Digital SLR camera and \textbf{Webcam (W)} contains images taken by web camera with different photo graphical settings. For all experiments, we use the source data with labels and target data without any labels for unsupervised domain adaptation. We conduct experiments on all six transfer tasks for all possible combinations of source and target pairs for the available three domains. The average performance of all transfer tasks are also calculated. 

\subsubsection{Office-Home}

The Office-Home dataset contains four domains and each domain
contains images from 65 different classes (categories). The four domains are \textbf{Art (Ar)}, \textbf{Clipart (Cl)}, \textbf{Product (Pr)} and \textbf{Real-World (Rw)}. Art domain contains the images from sketches, paintings, ornamentation form of artistic depictions of images. Clipart domain is the collection of clipart images. The images of Product domain have no background, and Real-World domain consists of images that are captured by a regular camera. It has around 15,500 images. Every category has an average of around 70 images and a maximum of 99 images. We conduct experiments on all 12 transfer tasks for all combinations of source and target pairs for the 4 domains. Figure \ref{fig:office} presents some sample images of 7 classes of Office-Home dataset.

\subsubsection{Office-Caltech}

The Office-Caltech is another popular benchmark dataset in the domain adaptation community which is formed by taking the 10 common classes shared by Office-31 and Caltech-256. It has four domains named \textbf{Amazon (A)}, \textbf{Webcam (W)}, \textbf{DSLR (D)} and \textbf{Caltech (C)}. We conduct experiments on all 12 transfer task as it has four different domains.

\subsection{Experimental Setup }

In our method we used two streams of Convolutional Neural Network (CNN). We extended AlexNet deep learning architecture which was pretrained on the ImageNet dataset for both stream of CNN. The dimension of the last fully connected layer (fc8) is set to the number of classes of the objects (31 for office 31, 65 for home-office and 10 for Office-Caltech datasets). We set the learning rate to 0.0001 to optimize the network. We set the batch size to 128, momentum to 0.9 and weight decay to $5 \times 10^{-4}$  during training phase.

%We conduct our experiment with one NVIDIA GeForce GTX 1070 GPU. The implementation is in  \textbf{Caffe} \cite{jia2014caffe} and our code will be made publicly available. 

\subsection{Results and Discussion}

In this section we provide the details of the performance of our method in the context of unsupervised domain adaptation where we use the labeled source data and unlabeled target data. Our proposed approach is compared with both conventional DA and  recently published deep architecture based approaches: Geodesic Flow Kernel (GFK) \cite{DBLP:conf/cvpr/GongSSG12}, Transfer Component Analysis (TCA) \cite{Pan:2009:DAV:1661445.1661635}, AlexNet (No adaptation) \cite{NIPS2012_4824}, VGG16 (No Adaptation) \cite{DBLP:journals/corr/SimonyanZ14a}, Domain Adversarial Neural Network (DANN) \cite{Ganin:2016:DTN:2946645.2946704}, Deep Correlation alignment (D-CORAL) \cite{dcoral}, DAN \cite{DBLP:journals/corr/Long015}, Deep Reconstruction-Classification Networks (DRCN) \cite{Ghifary2016}, Residual
Transfer Networks (RTN) \cite{DBLP:journals/corr/Long0J16}, and Deep Hashing Network (DAH) \cite{venkateswara2017Deep}.

\restylefloat{table}
\begin{table*}[!htbp]
\fontsize{8}{16}\selectfont 
\begin{center}
\begin{tabular}{l||c|c|c|c|c|c||cc}
\rowcolor[gray]{0.95}
\hline

%\textbf{Methods} & \textbf{$A\rightarrow$W} &
\textbf{Methods} & \textbf{A-W} & \textbf{D-W} & \textbf{D-A} & \textbf{W-A}& \textbf{W-D} & \textbf{A-D} & \textbf{Avg.}\\
\hline\hline

TCA \cite{5640675} &21.5 &50.1  &8.0 &14.6 &58.4 &11.4 &27.3\\
GFK \cite{DBLP:conf/cvpr/GongSSG12} &19.7 &49.7  &7.9  &15.8 &63.1 &10.6 &27.8\\

VGG16 \cite{DBLP:journals/corr/SimonyanZ14a} &63.9 &81.6 &46.9 &\textbf{54.1} &91.9 &63.1 &66.9\\
AlexNet \cite{NIPS2012_4824} &53.4 &79.9 &46.9 &47.5 &84.1 &55.6 &61.2\\
DANN \cite{Ganin:2016:DTN:2946645.2946704}  &\textbf{73.9 }&94.9  &-  &- &99.5 &- &-\\

%&73.9 &94.9 &99.5 &73.8 &69.8  &53.8&77.6\\

D-CORAL \cite{dcoral}  &67.2 &94.5  &52.6  &51.6 &98.7 &64.9 &71.6\\
DAN \cite{DBLP:journals/corr/Long015}  &68.5 &96.0  &50.0  &49.8 &\textbf{99.0} &66.8  &71.7\\
%WMMD \cite{DBLP:journals/corr/YanDLWXZ17}  &66.8 &95.9 &98.7 &64.5 &53.8  &52.7&72.1\\%
%DDC \cite{DBLP:journals/corr/TzengHZSD14}  &61.5 &95.3 &98.5 &64.9 &47.2 &49.4&69.5\\%

DRCN \cite{Ghifary2016} &68.7 &96.4  &\textbf{56.0}  &54.9 &99.0 &66.8  &73.6\\

RTN \cite{DBLP:journals/corr/Long0J16}  &73.3 &96.8  &50.5  &51.0 &99.6 &71.0   &73.7\\

DAH \cite{venkateswara2017Deep}  &68.3 &96.1  &55.5  &53.0 &98.8 &66.5  &73.0\\
\hline
\hline
\textbf{Our method}  &72.1 &\textbf{97.3}  &54.6  &53.9 &98.7 &\textbf{71.2}  &\textbf{74.6}\\
\hline
\end{tabular}
\end{center}
\caption{Image classification accuracies for deep domain adaptation on the Office-31 dataset.We use the standard protocol for unsupervised domain adaptation where source data are labeled, but target data are unlabeled. A - W indicates A (Amazon) is source and W (Webcam) is target.}
\label{office31}
\end{table*}

\begin{table*} [!htbp]
\fontsize{8}{20}\selectfont 
\begin{center}
\begin{tabular}{l||c|c|c|c|c|c|c|c|c|c|c|c||cc}
\rowcolor[gray]{0.95}
\hline

\textbf{Methods} & \textbf{A-C} & \textbf{A-P } & \textbf{A-R} & \textbf{C-A} & \textbf{C-P} & \textbf{C-R}& \textbf{P-A}& \textbf{P-C}& \textbf{P-R}& \textbf{R-A}& \textbf{R-C}& \textbf{R - P}&  \textbf{Avg.}\\
\hline\hline

TCA \cite{5640675} &19.93 &32.08 &35.71 &19.00 &31.36 &31.74 &21.92 &23.64 &42.12 &30.74 &27.15 &48.68 &30.34\\
GFK \cite{DBLP:conf/cvpr/GongSSG12} &21.60 &31.72 &38.83 &21.63 &34.94  &34.20&24.52 &25.73 &42.92 &32.88 &28.96 &50.89 &32.40\\

  VGG16 \cite{DBLP:journals/corr/SimonyanZ14a}  &30.40 &\textbf{45.92} &\textbf{57.54} &35.40 &48.67  &50.75 &\textbf{35.77} &30.51 &60.20 &49.62 &34.54 &64.00  &45.28\\
AlexNet \cite{NIPS2012_4824}  &27.40 &34.53 &45.04 &32.40 &43.90  &46.72 &29.76 &32.94 &50.20 &40.74 &35.07 &55.99  &39.74\\
DANN \cite{Ganin:2016:DTN:2946645.2946704}  &33.33 &42.96 &54.42 &32.26 &49.13  &49.76 &30.49& \textbf{ 38.14} &56.76 &44.71 &42.66 &64.65  &44.94\\

D-CORAL \cite{dcoral}  &32.18 &40.47 &54.45 &31.47 &45.8  &47.29 &30.03 &32.33 &55.27 &44.73 &42.75 &59.40  &42.79\\
DAN \cite{DBLP:journals/corr/Long015}  &30.66 &42.17 &54.13 &32.83 &47.59  &49.78 &29.07 &34.05 &56.70 &43.58 &38.25&62.73  &43.46\\

RTN \cite{DBLP:journals/corr/Long0J16} &31.23 &40.19 &54.56 &32.46 &46.60  &48.25 &28.20 &32.89 &56.38 &45.53 &44.74 &61.28  &43.53\\

DAH \cite{venkateswara2017Deep}  &31.64 &40.75 &51.73 &34.69 &51.93  &52.79 &29.91 &39.63 &60.71 &44.99 &45.13 &62.54  &45.54\\

\hline
\hline
%\textbf{2 MMD} &32.05 &40.75 &51.73 &34.69 &51.93  &52.79 &29.91 &39.63 &60.71 &44.99 &45.13 &62.54  &45.54\\%
%\textbf{2 CORAL} &33.24 &42.05 &55.57 &33.12 &47.25  &49.68 &32.52 &34.85 &57.98 &47.89 &43.05 &63.29  &-\\%
\textbf{Our method}  &\textbf{35.15} &44.35 &57.17 &\textbf{36.82} &\textbf{52.45}  &\textbf{53.67}&34.80 &37.17 &\textbf{62.15} &\textbf{49.95}&\textbf{46.29} &\textbf{66.05} &\textbf{48.00}\\
\hline
\end{tabular}
\end{center}
\caption{Image classification accuracies for deep domain adaptation on the Office-Home dataset. We use the standard protocol for unsupervised domain adaptation where source data are labeled, but target data are unlabeled. Ar - Cl indicates Ar (Art) is source domain and Cl (Clipart) is target domain.}
\label{office-home}
\end{table*}

\begin{table*} [!htbp]
\fontsize{9}{20}\selectfont 
\begin{center}
\begin{tabular}{l||c|c|c|c|c|c|c|c|c|c|c|c||cc}
\rowcolor[gray]{0.95}
\hline

\textbf{Methods} & \textbf{A-W} & \textbf{D-W } &  \textbf{D-A} & \textbf{W-A}& \textbf{W-D} & \textbf{A-D} & \textbf{A-C}& \textbf{W-C}&  \textbf{C-W}& \textbf{C-D}&  \textbf{D-C}& \textbf{C-A}& \textbf{Avg.}\\
\hline\hline

TCA \cite{5640675} &84.4 &96.9  &90.4 &85.6 &99.4 &82.8 &81.2 &75.5  &88.1 &87.9 &79.6 &92.1  &87.0\\
GFK \cite{DBLP:conf/cvpr/GongSSG12} &89.5 &97.0  &89.8  &88.5  &98.1 &86.0 &76.2 &77.1  &78.0 &77.1  &77.9 &90.7 &85.5\\

 % VGG16 \cite{DBLP:journals/corr/SimonyanZ14a}  &- &\textbf{-} &\textbf{-} &- &-  &- &\textbf{-} &- &- &- &- &-  &-\\
AlexNet \cite{NIPS2012_4824}  &79.5 &97.7  &87.1  &83.8 &\textbf{100.0} &87.4  &83.0 &73.0  &83.7 &87.1 &79.0 &91.9 &86.1\\
%DANN \cite{Ganin:2016:DTN:2946645.2946704}  &- &- &- &- &-  &- &-& \textbf{ -} &- &- &- &-  &-\\

D-CORAL \cite{dcoral}  &89.8 &97.3  &91.0  &91.9 &\textbf{100.0} &90.5 &83.7 &81.5 &90.1 &88.6 &80.1 &92.3  &89.7\\

DAN \cite{DBLP:journals/corr/Long015}  &91.8 &98.5  &90.0  &92.1 &\textbf{100.0} &91.7  &84.1 &81.2  &90.3 &89.3 &80.3 &92.0  &90.1\\

RTN \cite{DBLP:journals/corr/Long0J16} &95.2 &99.2  &93.8  &92.5 &\textbf{100.0} &95.5 &88.1 &\textbf{86.6}  &\textbf{96.9} &\textbf{94.2} &84.6 & \textbf{93.7}  &93.4\\

%DAH \cite{venkateswara2017Deep}  &- &- &- &- &-  &- &- &- &- &- &- &-  &-\\

\hline
\hline
%\textbf{2 MMD} &32.05 &40.75 &51.73 &34.69 &51.93  &52.79 &29.91 &39.63 &60.71 &44.99 &45.13 &62.54  &45.54\\%
%\textbf{2 CORAL} &33.24 &42.05 &55.57 &33.12 &47.25  &49.68 &32.52 &34.85 &57.98 &47.89 &43.05 &63.29  &-\\%
\textbf{Our method}  &\textbf{95.7} &\textbf{99.4}  &\textbf{94.7}  &\textbf{94.8} &\textbf{100.0} &\textbf{96.6} &\textbf{89.1} &86.5  &95.2 &93.4 &\textbf{84.7} &93.6 &\textbf{93.6}\\
\hline
\end{tabular}
\end{center}
\caption{Image classification accuracies for deep domain adaptation on the Office-Caltech dataset. We use the protocol for unsupervised domain adaptation where source data are labeled, but target data are unlabeled. A - C indicates A (Amazon) is source and C (Caltech) is target.}
\label{caltech}
\end{table*}

TCA is a traditional domain adaptation approach based on MMD-regularized Kernel primary component analysis (PCA). GFK is a subspace based domain adaptation approach. Both TCA and GFK do not use a deep neural architecture. These methods are not end-to-end approach. At first features are extracted and then the features are used in domain adaptation networks. Both AlexNet and VGG16 deep convolutional neural networks are also used as deep feature extractors without adaptation techniques to show that a standalone deep architecture works better than conventional domain adaptation techniques. DANN introduces a deep learning approach domain adaptation technique by integrating a gradient reversal layer into the standard architecture. D-CORAL is also another deep domain adaptation architecture where second order statistics alignment technique is used to adapt features. DAN uses MMD to minimize the dissimilarity between source and target domains. DRCN introduces an unsupervised domain adaptation model which reconstruct source images that have a similar appearance to or qualities in common with the target images. RTN introduces residual transfer network where classifiers and features are adapted simultaneously. DAH uses deep hashing network for unsupervised domain adaptation. In DAH, MMD is utilized to decrease the dissimilarities between the source and target domains.

We use Caffe \cite{jia2014caffe} framework to implement our proposed method. We use Alexnet  architecture \cite{NIPS2012_4824}. We conduct experiments with one NVIDIA GeForce GTX 1070 Graphics Processing Unit (GPU). For unsupervised domain adaptation techniques, we follow the standard protocol where the source data are labeled, but the target data are unlabeled. We make a comparison based on average classification accuracy for each transfer task.

As shown in Table \ref{office31}, \ref{office-home} and \ref{caltech}, we compare the results of our proposed method with state-of-the-art approaches on
three datasets (Office 31, Office-Home and Office-Caltech) in the context of classification accuracy. The classification accuracy of a model $A_i$ depends on the images correctly identified. We evaluated all the methods by using the following formula:

\begin{equation}
{A_i}  = \frac{t}{n} \times {100},
\end{equation}

where, t is the total number of correctly classified images, and n belong to the total images.

For Office-31 dataset, we report the image classification results in Table \ref{office31} for target data on different transfer tasks. In Table \ref{office-home}, the target data classification accuracy are reported for Office-Home dataset on twelve transfer tasks. For Office-Caltech dataset, the target classification accuracy on different transfer tasks are reported in Table \ref{caltech}. The accuracies stand for the percentage of correctly classified target images.

For Office-31 dataset, the previous best average result achieved by \cite{DBLP:journals/corr/Long0J16} and \cite{Ghifary2016} which are 73.7\% and 73.6\% respectively. In contrast with their approach, our combined CORAL and MMD loss outperforms their results by  0.9\% and 1.0\% respectively. For Office-Home dataset, our proposed
method achieves average 48.0\% classification accuracy which outperforms most state-of-the-art approaches, such as, DAH \cite{venkateswara2017Deep} by 2.46\%. For the Office-Caltech dataset, the existing best result was achieved by \cite{DBLP:journals/corr/Long0J16}. Our proposed method beats their average classification accuracy by 0.2\%. Thus, the proposed model based on MMD and CORAL outperforms all comparison methods on most transfer tasks on the datasets. From Table \ref{office31}, \ref{office-home} and \ref{caltech}, we can see that the proposed method achieves better average performance than other baseline conventional and deep domain adaptation methods.

These results provides the suggestion that our proposed method is capable to acquire better classifiers which are adaptive in between domains and transferable features to solve domain adaptation issue.

From all the results in terms of image classification, we can find the following observations:

\begin{itemize}
\item Traditional deep learning approaches without domain adaptation perform better than the standard domain adaptation methods.

\item The proposed unsupervised deep domain adaptation based on joint aligning of the second order statistics and maximum mean discrepancy outperforms the state-of-the-art methods.  

\item Our models works better where the number of classes of objects are more. For an example Office-Home dataset contains 65 categories and we achieved $48\%$ accuracy using our model.

\end{itemize}

\subsection{Visualization}

\begin{figure*}
\begin{center}
\includegraphics[width=1.0\linewidth]{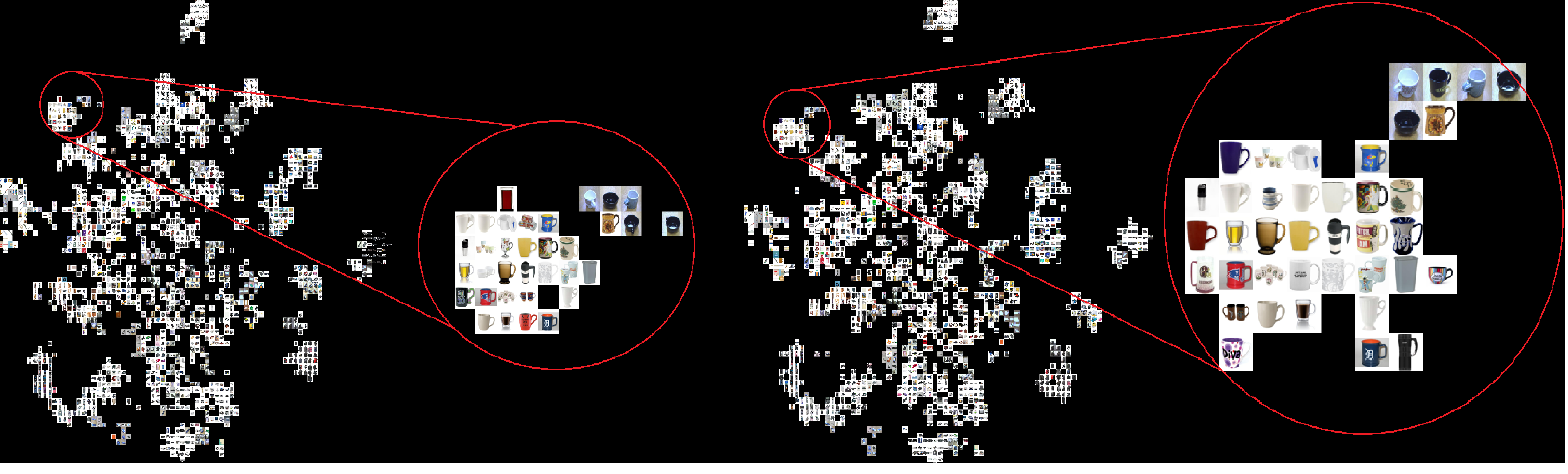}
\end{center}
   \caption{t-SNE embedding of images that are taken from Amazon and Webcam domains using AlexNet model (left) and using two MMD and CORAL metrics in between fc7 and fc8 layers of the two stream CNN (right). While mixing domains, It is observed that the clusters created by our proposed model that can separate classes much more efficiently than AlexNet model where there is no domain adaptation technique is applied.}
\label{fig:tsne}
\end{figure*}

We use t-SNE for embedding visualization. To produce an embedding, we take images from Amazon and Webcam domains of Office-31 dataset. We use the CNN model to acquire the corresponding fc7-4096 dimensional vector for each image. After that, we plug these fc7-4096 vectors into t-SNE and generate 2-dimensional vector for each image. We plot a t-SNE embedding in Figure \ref{fig:tsne} of images that are taken from Amazon and Webcam domains using our learned representation (right) and make a comparison it to an embedding formed with AlexNet in Figure \ref{fig:tsne} (left). Examining the embeddings, we found that the clusters created by our model separate the classes while mixing the domains much more efficiently than the AlexNet approach where there is no domain adaptation technique is applied.

\section{Conclusion}

In this paper, we introduce an unsupervised deep domain adaptation architecture where the features and classifiers are adapted jointly. The source and target features are adapted by aligning covariances as well as maximum mean discrepancy and the classifiers are adapted by minimizing the entropy loss of the target data. Extensive Experimental results on standard benchmark datasets suggest the state-of-the art performance. Prior deep domain adaptation techniques either use MMD or CORAL to decrease the mismatch between the source and target data. However, unlike previous work, we use both MMD and CORAL to adapt the features across domains. This makes our method a decent supplement to existing procedures.

\bibliographystyle{ieee}
\bibliography{samplepaper}

\begin{thebibliography}{10}\itemsep=-1pt

\bibitem{Choi_2018_CVPR}
Y.~Choi, M.~Choi, M.~Kim, J.-W. Ha, S.~Kim, and J.~Choo.
\newblock Stargan: Unified generative adversarial networks for multi-domain
  image-to-image translation.
\newblock In {\em The IEEE Conference on Computer Vision and Pattern
  Recognition (CVPR)}, 2018.

\bibitem{6639346}
G.~E. Dahl, T.~N. Sainath, and G.~E. Hinton.
\newblock Improving deep neural networks for lvcsr using rectified linear units
  and dropout.
\newblock In {\em International Conference on Acoustics, Speech and Signal
  Processing (ICASSP)}, 2013.

\bibitem{Donahue:2014:DDC:3044805.3044879}
J.~Donahue, Y.~Jia, O.~Vinyals, J.~Hoffman, N.~Zhang, E.~Tzeng, and T.~Darrell.
\newblock Decaf: A deep convolutional activation feature for generic visual
  recognition.
\newblock In {\em International Conference on Machine Learning (ICML)}, 2014.

\bibitem{6751479}
B.~Fernando, A.~Habrard, M.~Sebban, and T.~Tuytelaars.
\newblock Unsupervised visual domain adaptation using subspace alignment.
\newblock In {\em The IEEE Conference on Computer Vision and Pattern
  Recognition (CVPR)}, 2013.

\bibitem{Ganin:2016:DTN:2946645.2946704}
Y.~Ganin, E.~Ustinova, H.~Ajakan, P.~Germain, H.~Larochelle, F.~Laviolette,
  M.~Marchand, and V.~Lempitsky.
\newblock Domain-adversarial training of neural networks.
\newblock {\em Journal of Machine Learning Research}, 17(1), 2016.

\bibitem{Ghifary2016}
M.~Ghifary, W.~B. Kleijn, M.~Zhang, D.~Balduzzi, and W.~Li.
\newblock {\em Deep Reconstruction-Classification Networks for Unsupervised
  Domain Adaptation}.
\newblock Springer International Publishing, Cham, 2016.

\bibitem{Girshick:2014:RFH:2679600.2679851}
R.~Girshick, J.~Donahue, T.~Darrell, and J.~Malik.
\newblock Rich feature hierarchies for accurate object detection and semantic
  segmentation.
\newblock In {\em The IEEE Conference on Computer Vision and Pattern
  Recognition (CVPR)}, 2014.

\bibitem{DBLP:conf/cvpr/GongSSG12}
B.~Gong, Y.~Shi, F.~Sha, and K.~Grauman.
\newblock Geodesic flow kernel for unsupervised domain adaptation.
\newblock In {\em The IEEE Conference on Computer Vision and Pattern
  Recognition (CVPR)}, 2012.

\bibitem{DBLP:journals/corr/GoodfellowBIAS13}
I.~Goodfellow, Y.~Bulatov, J.~Ibarz, S.~Arnoud, and V.~Shet.
\newblock Multi-digit number recognition from street view imagery using deep
  convolutional neural networks.
\newblock In {\em International Conference on Learning Representations (ICLR)},
  2014.

\bibitem{Hong_2018_CVPR}
W.~Hong, Z.~Wang, M.~Yang, and J.~Yuan.
\newblock Conditional generative adversarial network for structured domain
  adaptation.
\newblock In {\em The IEEE Conference on Computer Vision and Pattern
  Recognition (CVPR)}, 2018.

\bibitem{Hu_2018_CVPR}
L.~Hu, M.~Kan, S.~Shan, and X.~Chen.
\newblock Duplex generative adversarial network for unsupervised domain
  adaptation.
\newblock In {\em The IEEE Conference on Computer Vision and Pattern
  Recognition (CVPR)}, 2018.

\bibitem{jia2014caffe}
Y.~Jia, E.~Shelhamer, J.~Donahue, S.~Karayev, J.~Long, R.~Girshick,
  S.~Guadarrama, and T.~Darrell.
\newblock Caffe: Convolutional architecture for fast feature embedding.
\newblock In {\em ACM International Conference on Multimedia}, 2014.

\bibitem{8354250}
A.~Khatun, S.~Denman, S.~Sridharan, and C.~Fookes.
\newblock A deep four-stream siamese convolutional neural network with joint
  verification and identification loss for person re-detection.
\newblock In {\em IEEE Winter Conference on Applications of Computer Vision
  (WACV)}, 2018.

\bibitem{DBLP:journals/corr/KoniuszTP16}
P.~Koniusz, Y.~Tas, and F.~Porikli.
\newblock Domain adaptation by mixture of alignments of second-or higher-order
  scatter tensors.
\newblock In {\em The IEEE Conference on Computer Vision and Pattern
  Recognition (CVPR)}, 2017.

\bibitem{NIPS2012_4824}
A.~Krizhevsky, I.~Sutskever, and G.~E. Hinton.
\newblock Imagenet classification with deep convolutional neural networks.
\newblock In {\em Neural Information Processing Systems (NIPS)}. 2012.

\bibitem{0483bd9444a348c8b59d54a190839ec9}
Y.~Lecun, Y.~Bengio, and G.~Hinton.
\newblock Deep learning.
\newblock {\em Nature}, 521(7553):436--444, 5 2015.

\bibitem{DBLP:journals/corr/Long015}
M.~Long, Y.~Cao, J.~Wang, and M.~I. Jordan.
\newblock Learning transferable features with deep adaptation networks.
\newblock In {\em International Conference on Machine Learning (ICML)}, 2015.

\bibitem{DBLP:journals/corr/Long0J16}
M.~Long, H.~Zhu, J.~Wang, and M.~I. Jordan.
\newblock Unsupervised domain adaptation with residual transfer networks.
\newblock In {\em Neural Information Processing Systems (NIPS)}, 2016.

\bibitem{DBLP:journals/corr/Long0J16a}
M.~Long, H.~Zhu, J.~Wang, and M.~I. Jordan.
\newblock Deep transfer learning with joint adaptation networks.
\newblock In {\em International Conference on Machine Learning (ICML)}, 2017.

\bibitem{Mancini_2018_CVPR}
M.~Mancini, L.~Porzi, S.~Rota~Bulò, B.~Caputo, and E.~Ricci.
\newblock Boosting domain adaptation by discovering latent domains.
\newblock In {\em The IEEE Conference on Computer Vision and Pattern
  Recognition (CVPR)}, 2018.

\bibitem{Murez_2018_CVPR}
Z.~Murez, S.~Kolouri, D.~Kriegman, R.~Ramamoorthi, and K.~Kim.
\newblock Image to image translation for domain adaptation.
\newblock In {\em The IEEE Conference on Computer Vision and Pattern
  Recognition (CVPR)}, 2018.

\bibitem{Pan:2009:DAV:1661445.1661635}
S.~J. Pan, I.~W. Tsang, J.~T. Kwok, and Q.~Yang.
\newblock Domain adaptation via transfer component analysis.
\newblock In {\em International Joint Conference on Artificial Intelligence
  (IJCAI)}, 2009.

\bibitem{5640675}
S.~J. Pan, I.~W. Tsang, J.~T. Kwok, and Q.~Yang.
\newblock Domain adaptation via transfer component analysis.
\newblock {\em IEEE Transactions on Neural Networks}, 22(2):199--210, Feb 2011.

\bibitem{5288526}
S.~J. Pan and Q.~Yang.
\newblock A survey on transfer learning.
\newblock {\em IEEE Transactions on Knowledge and Data Engineering},
  22(10):1345--1359, Oct 2010.

\bibitem{7078994}
V.~M. Patel, R.~Gopalan, R.~Li, and R.~Chellappa.
\newblock Visual domain adaptation: A survey of recent advances.
\newblock {\em IEEE Signal Processing Magazine}, 32(3), 2015.

\bibitem{Pinheiro_2018_CVPR}
P.~O. Pinheiro.
\newblock Unsupervised domain adaptation with similarity learning.
\newblock In {\em The IEEE Conference on Computer Vision and Pattern
  Recognition (CVPR)}, 2018.

\bibitem{Saenko:2010:AVC:1888089.1888106}
K.~Saenko, B.~Kulis, M.~Fritz, and T.~Darrell.
\newblock Adapting visual category models to new domains.
\newblock In {\em European Conference on Computer Vision (ECCV)}, 2010.

\bibitem{Saito_2018_CVPR}
K.~Saito, K.~Watanabe, Y.~Ushiku, and T.~Harada.
\newblock Maximum classifier discrepancy for unsupervised domain adaptation.
\newblock In {\em The IEEE Conference on Computer Vision and Pattern
  Recognition (CVPR)}, 2018.

\bibitem{DBLP:journals/corr/SimonyanZ14a}
K.~Simonyan and A.~Zisserman.
\newblock Very deep convolutional networks for large-scale image recognition.
\newblock {\em CoRR}, abs/1409.1556, 2014.

\bibitem{coral}
B.~Sun, J.~Feng, and K.~Saenko.
\newblock Return of frustratingly easy domain adaptation.
\newblock In {\em AAAI Conference on Artificial Intelligence}, 2016.

\bibitem{DBLP:journals/corr/SunFS16}
B.~Sun, J.~Feng, and K.~Saenko.
\newblock Correlation alignment for unsupervised domain adaptation.
\newblock In {\em Domain Adaptation in Computer Vision Applications.}, pages
  153--171. 2017.

\bibitem{dcoral}
B.~Sun and K.~Saenko.
\newblock Deep coral: Correlation alignment for deep domain adaptation.
\newblock In {\em European Conference on Computer Vision Workshops}, 2016.

\bibitem{DBLP:journals/corr/SutskeverVL14}
I.~Sutskever, O.~Vinyals, and Q.~V. Le.
\newblock Sequence to sequence learning with neural networks.
\newblock In {\em Neural Information Processing Systems (NIPS)}, 2014.

\bibitem{5995347}
A.~Torralba and A.~A. Efros.
\newblock Unbiased look at dataset bias.
\newblock In {\em The IEEE Conference on Computer Vision and Pattern
  Recognition (CVPR)}, 2011.

\bibitem{DBLP:journals/corr/TzengHDS15}
E.~Tzeng, J.~Hoffman, T.~Darrell, and K.~Saenko.
\newblock Simultaneous deep transfer across domains and tasks.
\newblock In {\em The IEEE Conference on International Conference on Computer
  Vision (ICCV)}, 2015.

\bibitem{DBLP:journals/corr/TzengHZSD14}
E.~Tzeng, J.~Hoffman, N.~Zhang, K.~Saenko, and T.~Darrell.
\newblock Deep domain confusion: Maximizing for domain invariance.
\newblock {\em CoRR}, abs/1412.3474, 2014.

\bibitem{venkateswara2017Deep}
H.~Venkateswara, J.~Eusebio, S.~Chakraborty, and S.~Panchanathan.
\newblock Deep hashing network for unsupervised domain adaptation.
\newblock In {\em The IEEE Conference on Computer Vision and Pattern
  Recognition (CVPR)}, 2017.

\bibitem{Volpi_2018_CVPR}
R.~Volpi, P.~Morerio, S.~Savarese, and V.~Murino.
\newblock Adversarial feature augmentation for unsupervised domain adaptation.
\newblock In {\em The IEEE Conference on Computer Vision and Pattern
  Recognition (CVPR)}, 2018.

\bibitem{DBLP:journals/corr/YanDLWXZ17}
H.~Yan, Y.~Ding, P.~Li, Q.~Wang, Y.~Xu, and W.~Zuo.
\newblock Mind the class weight bias: Weighted maximum mean discrepancy for
  unsupervised domain adaptation.
\newblock In {\em The IEEE Conference on Computer Vision and Pattern
  Recognition (CVPR)}, 2017.

\bibitem{YANG2018615}
B.~Yang, A.~J. Ma, and P.~C. Yuen.
\newblock Learning domain-shared group-sparse representation for unsupervised
  domain adaptation.
\newblock {\em Pattern Recognition}, 81:615 -- 632, 2018.

\end{thebibliography}

\end{document}